\documentclass[twocolumn,letterpaper]{article} % DO NOT CHANGE THIS
\usepackage{aaai24}  % DO NOT CHANGE THIS
\usepackage{times}  % DO NOT CHANGE THIS
\usepackage{helvet}  % DO NOT CHANGE THIS
\usepackage{courier}  % DO NOT CHANGE THIS
\usepackage[hyphens]{url}  % DO NOT CHANGE THIS
\usepackage{graphicx} % DO NOT CHANGE THIS
\usepackage{natbib}  % DO NOT CHANGE THIS AND DO NOT ADD ANY OPTIONS TO IT
\usepackage{caption} % DO NOT CHANGE THIS AND DO NOT ADD ANY OPTIONS TO IT
\usepackage{algorithm}
\usepackage{algorithmic}
\usepackage{amssymb}
\usepackage{amsmath}
\usepackage{booktabs}
\graphicspath{{./images/}}

\usepackage{etoolbox}%% cuizhen
\newcommand{\MyMapTemplatePrefix}[4]{\expandafter#1\csname#3#4\endcsname{#2{#4}}}
\newcommand{\MyMapTemplatePrefixNew}[5]{\expandafter#1\csname#4#5\endcsname{#2{#3{#5}}}}
\forcsvlist{\MyMapTemplatePrefix {\def} {\mathbf} {}} {A,B,C,D,E,F,G,H,I,J,K,L,M,N,O,P,Q,R,S,T,U,V,W,X,Y,Z}
\forcsvlist{\MyMapTemplatePrefix {\def} {\mathbf} {}} {a,b,c,d,e,f,g,h,i,j,k,l,m,n,o,p,q,r,s,t,u,v,w,x,y,z,1,0}
\forcsvlist{\MyMapTemplatePrefix {\def} {\widetilde} {wt}} {A,B,C,D,E,F,G,H,I,J,K,L,M,N,O,P,Q,R,S,T,U,V,W,X,Y,Z}
\forcsvlist{\MyMapTemplatePrefix {\def} {\widetilde} {wt}} {a,b,c,d,e,f,g,h,i,j,k,l,m,n,o,p,q,r,s,t,u,v,w,x,y,z} % r
\forcsvlist{\MyMapTemplatePrefixNew {\def} {\widetilde}{\mathbf} {tb}} {A,B,C,D,E,F,G,H,I,J,K,L,M,N,O,P,Q,R,S,T,U,V,W,X,Y,Z}
\forcsvlist{\MyMapTemplatePrefixNew {\def} {\widetilde}{\mathbf} {tb}} {a,b,c,d,e,f,g,h,i,j,k,l,m,n,o,p,q,r,s,t,u,v,w,x,y,z}
\forcsvlist{\MyMapTemplatePrefixNew {\def} {\overleftarrow}{\mathbf}{ol}} {A,B,C,D,E,F,G,H,I,J,K,L,M,N,O,P,Q,R,S,T,U,V,W,X,Y,Z}
\forcsvlist{\MyMapTemplatePrefixNew {\def}{\overrightarrow}{\mathbf}{or}} {A,B,C,D,E,F,G,H,I,J,K,L,M,N,O,P,Q,R,S,T,U,V,W,X,Y,Z}
\forcsvlist{\MyMapTemplatePrefix {\def} {\widehat} {wh}} {A,B,C,D,E,F,G,H,I,J,K,L,M,N,O,P,Q,R,S,T,U,V,W,X,Y,Z}
\forcsvlist{\MyMapTemplatePrefix {\def} {\widehat} {wh}} {a,b,c,d,e,f,g,h,i,j,k,l,m,n,o,p,q,r,s,t,u,v,w,x,y,z}
\forcsvlist{\MyMapTemplatePrefixNew {\def} {\widehat}{\mathbf} {hb}} {A,B,C,D,E,F,G,H,I,J,K,L,M,N,O,P,Q,R,S,T,U,V,W,X,Y,Z}
\forcsvlist{\MyMapTemplatePrefixNew {\def} {\widehat}{\mathbf} {hb}} {a,b,c,d,e,f,g,h,i,j,k,l,m,n,o,p,q,r,s,t,u,v,w,x,y,z}
\forcsvlist{\MyMapTemplatePrefix {\def} {\mathcal}{mc}} {A,B,C,D,E,F,G,H,I,J,K,L,M,N,O,P,Q,R,S,T,U,V,W,X,Y,Z}
\forcsvlist{\MyMapTemplatePrefix {\def} {\mathpzc}{mc}} {a,b,c,d,e,f,g,h,i,j,k,l,m,n,o,p,q,r,s,t,u,v,w,x,y,z}
\forcsvlist{\MyMapTemplatePrefixNew {\def} {\widetilde}{\mathcal} {tc}} {A,B,C,D,E,F,G,H,I,J,K,L,M,N,O,P,Q,R,S,T,U,V,W,X,Y,Z}
\forcsvlist{\MyMapTemplatePrefix {\def} {\mathbb} {mb}} {A,B,C,D,E,F,G,H,I,J,K,L,M,N,O,P,Q,R,S,T,U,V,W,X,Y,Z}
\forcsvlist{\MyMapTemplatePrefix {\DeclareMathOperator} {} {} } {tr,diag,sgn}

\usepackage{newfloat}
\usepackage{listings}
\DeclareCaptionStyle{ruled}{labelfont=normalfont,labelsep=colon,strut=off} % DO NOT CHANGE THIS
\floatstyle{ruled}
\newfloat{listing}{tb}{lst}{}
\floatname{listing}{Listing}

\title{Dual-Stream Diffusion Net for Text-to-Video Generation}
\author{
	Binhui Liu\textsuperscript{\rm 1}, Xin Liu\textsuperscript{\rm 2}, Anbo Dai\textsuperscript{\rm 2}, Zhiyong Zeng\textsuperscript{\rm 1}, \\
	Dan Wang\textsuperscript{\rm 1}, Zhen Cui\textsuperscript{\rm 1}, Jian Yang\textsuperscript{\rm 3}	
}
\affiliations{
    \textsuperscript{\rm 1}Nanjing University of Science and Technology, Nanjing, China\\
	\textsuperscript{\rm 2}SeetaCloud, Nanjing, China\\
	\textsuperscript{\rm 3}Nankai University, Tianjin, China\\
	\{lbhasura, zhiyong.zeng, zhen.cui\}@njust.edu.cn, \\
	\{xin.liu, anbo.dai\}@seetacloud.com, csjyang@nankai.edu.cn \\
    
}

\usepackage{bibentry}
\begin{document}

\twocolumn[{
\renewcommand\twocolumn[1][]{#1}
\maketitle
\begin{center}
    \captionsetup{type=figure}
    \includegraphics[width=1\textwidth]{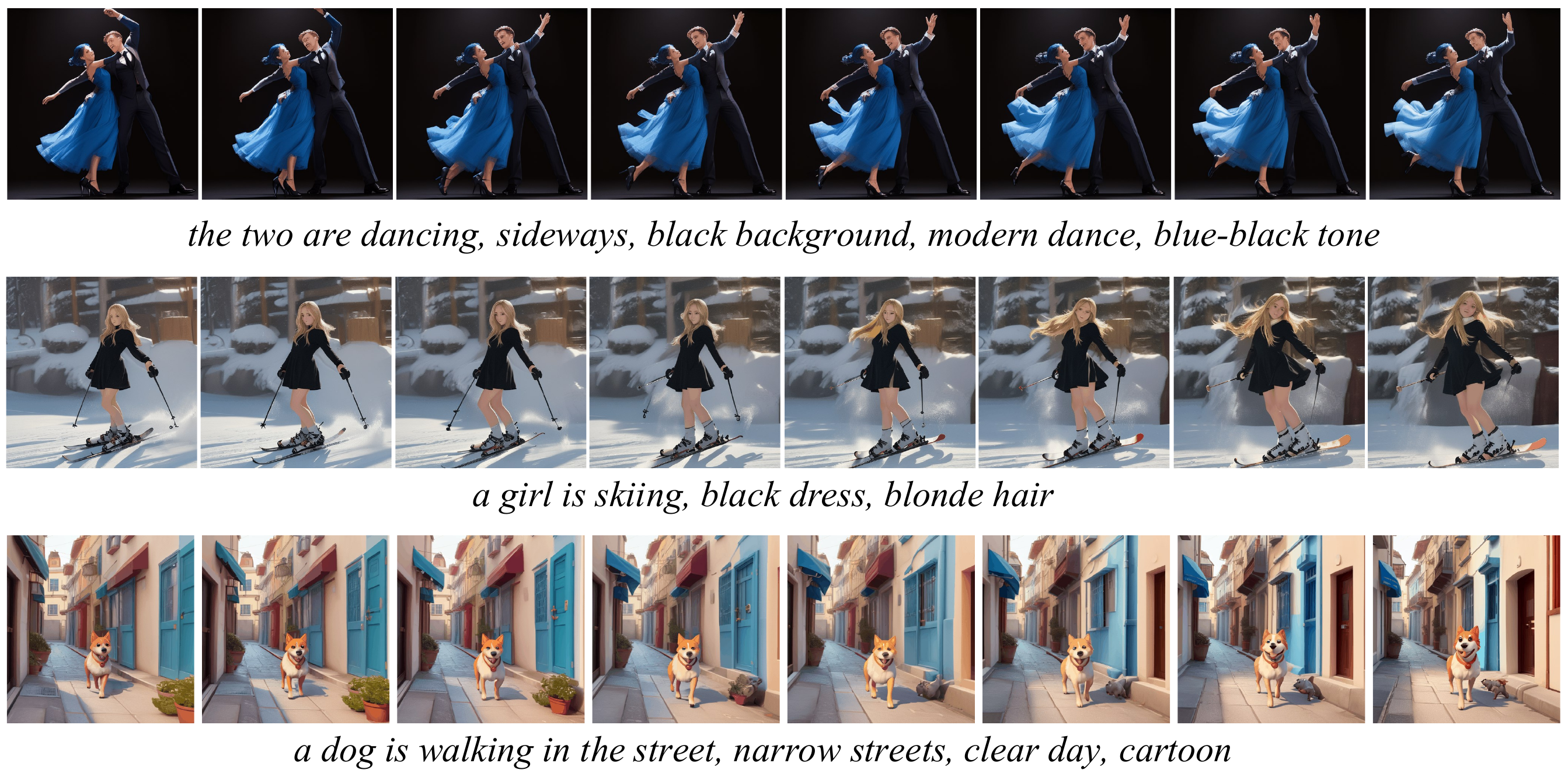}
    \captionof{figure}{Samples generated by our method.}
    \label{fig:show_figure}
\end{center}
}]

\begin{abstract}
With the emerging diffusion models, recently, text-to-video generation has aroused  increasing attention. But an important bottleneck therein is that generative videos often tend to carry some flickers and artifacts. In this work, we propose a dual-stream diffusion net (DSDN) to improve the consistency of content variations in generating videos. In particular, the designed two diffusion streams, video content and motion branches, could not only run separately in their private spaces for producing personalized video variations as well as content, but also be well-aligned between the content and motion domains through leveraging our designed cross-transformer interaction module, which would benefit the smoothness of generated videos. Besides, we also introduce motion decomposer and combiner to faciliate the operation on video motion. Qualitative and quantitative experiments demonstrate that our method could produce amazing continuous videos with fewer flickers (see Fig.~\ref{fig:show_figure})\footnote{Please see the videos in supplementary material, and more info including code could be found in the anonymous website: https://anonymous.4open.science/r/Private-C3E8}. 
\end{abstract}

\section{Introduction}

In the realm of artificial intelligence generated content, one of the most exciting and challenging tasks is the transformation from text into visual content. This task not only benefits for our understanding of natural language processing but also promotes computer vision techniques. Meantime, it will cause immense potential applications in entertainment, advertising, education, and surveillance. 
Over the past few years, there has been substantial progress in developing models that convert textual descriptions into images~\cite{alex2021glid, Aditya2022Hierarchical, Robin2022High}. In contrast to images, videos could carry/express richer content. For textual descriptions, videos can capture and convey intricate narratives well therein~\cite{Max2021Frozen, Lucas2020Short}. Recently, the text-to-video generation has been paid an increasing attention, specifically accompanying with the rise of diffusion models.

One critical challenge lies in reasonable and continuous content generation in spatial-temporal domain for videos, not only spatial content  learnt by image-based diffusion models. Presently, the existing methods~\cite{Andreas2023Align, Wenyi2022CogVideo, Uriel2022Make} of text-to-video generation primarily focused on reproducing visual content from text. Due to the insufficiency in modeling video dynamics, their generated videos often contain many flickers and are intermittent in visual effects. To address this issue, in this work, our goal is to increase the consistency of motion and content between these video frames, while augmenting the motion diversity of generated videos, so as to generate better visually continuous videos.  

To this end, here we propose a dual-stream diffusion net (DSDN) to boost the consistency of content variations in generating videos. To characterize the motion information of video, in particular, we introduce a motion branch to encode video content variations besides the branch of video content. Hereby, we construct a two-branch diffusion network, video motion stream as well as video content stream. To make full of large-scale image generation model, 
the content stream runs upon a pre-trained text-to-image conditional diffusion model, but meantime is updated incrementally with a parallel network for personalized video content generation. On the parallel step, the variations of video frames, i.e. motion, takes a separate probability diffusion process through employing 3D-UNet, so that personalize motion information could be generated. To align the generated content and motion, we design a dual-stream transformation interaction module by using cross-attention between the two streams. Accordingly, the motion stream is integrated with the content stream during the denoising process, which allows each stream to serve as contextual information for the other. Besides, we also introduce motion decomposer and combiner to faciliate the operation on motion. We conducted qualitative and quantitative experimental verification, and the experiments demonstrate that our method enable to produce better visually continuous videos, as shown in Fig.~\ref{fig:show_figure}.

In the end, we briefly summarize the contributions to the realm of text-to-video generation: i) propose a Dual-Stream Diffusion Net (DSDN) to enhance the consistency and diversity of generated videos, where the motion is specifically modeled as a single branch that distinguishes from most existing video diffusion methods; ii) design some useful modules, including personalized content/motion generation, dual-stream transformation interaction, to align content and motion while preserving the diversity of generated samples; iii) qualitative and quantitative evaluations demonstrate that DSDN could effectively generates videos with remarkable consistency and diversity.

\section{Related Work}
The development and evolution of models for converting textual descriptions into visual content have been a consistent focus in the field of artificial intelligence. The research has gradually transitioned from text-to-image models to more dynamic and complex text-to-video generation models.

\subsubsection{Text-to-Image Generation}
Early efforts were dedicated to developing techniques for text-to-image synthesis. 
The Denoising Diffusion Probabilistic Model (DDPM)~\cite{Jonathan2020Denoising, Jiaming2022Denoising} has garnered significant attention owing to its remarkable ability to generate high-quality images. This innovative model has exceeded the performance of previous generative adversarial networks (GANs)~\cite{Ian2020Generative}, setting a new benchmark in the field. Furthermore, the DDPM has a unique feature: it can be trained with text guidance, empowering users to generate images from textual inputs.
Several notable advancements have been made in this area. For instance, GLIDE~\cite{alex2021glid} adopts classifier-free guidance and trains the diffusion model using large-scale text-image pairs. DALLE-2~\cite{Aditya2022Hierarchical} uses CLIP~\cite{Alec2021Learning} latent space as a condition, which significantly enhances the performance. Imagen~\cite{Chitwan2022Photorealistic} employs a T5~\cite{Colin2020Exploring} coupled with cascaded diffusion models to generate high-resolution images. The Latent Diffusion Model (LDM)~\cite{Aditya2022Hierarchical} proposes forwarding the diffusion process in latent space, demonstrating higher efficiency than other diffusion models.
%

% Text to Video
\subsubsection{Text-to-Video Generation}
Despite these advancements in text-to-image models, transitioning to text-to-video synthesis presented new challenges, mainly due to the temporal dependencies between video frames and the need to maintain motion semantics throughout the video sequence. 
Early works in this regard include GAN-based methods~\cite{Carl2016Generating, Aidan2019Adversarial} and auto-regressive one~\cite{Nal2017Video, Wenyi2022CogVideo}.
In the context of unconditional video generation, Ho et al.~\cite{Jonathan2022Video} successfully extended the DDPM models initially designed for images into the video domain, leading to the development of a 3D U-Net architecture. Harvey et al.~\cite{William2022Flexible} put forth an innovative approach wherein they modeled the distribution of subsequent video frames in an auto-regressive manner.
Our primary focus, however, lies in synthesizing videos in a controllable manner - more specifically, in text-conditional video generation. Exploring this avenue, Hong et al.~\cite{Wenyi2022CogVideo} proposed CogVideo, an autoregressive framework that models the video sequence by conditioning it on the given text and the previous frames. Similarly, Levon et al.~\cite{Levon2023Zero} proposed the Text2Video-Zero, a text-to-video generation method based on the text-to-image model stable diffusion~\cite{Robin2022High}, which can not only directly generate text-to-video, but also directly complete image editing tasks.
The current issue in the text-to-video domain is that generative videos often tend to carry some flickers and artifacts. 
Few attempts made to capture both the visual and dynamic aspects of videos include the latent stream diffusion models proposed by Ni et al.~\cite{Haomiao2023Conditional}, and et al.~\cite{Sihyun2023Video} projected latent video diffusion model for generating long video through the integration of spatial and temporal information flow.
These have been successfully used in tasks such as generating high-quality images from textual descriptions~\cite{alex2021glid, Aditya2022Hierarchical, Robin2022High}, while their potential in generating dynamic videos from text remains largely untapped.
Our work is inspired by these previous research efforts and seeks to address the pitfalls common in existing models. We introduce a novel dual-stream diffusion net to improve the consistency of content variations in generating videos.

\begin{figure*}[!thp]
	\centering
	\includegraphics[width=0.90\textwidth]{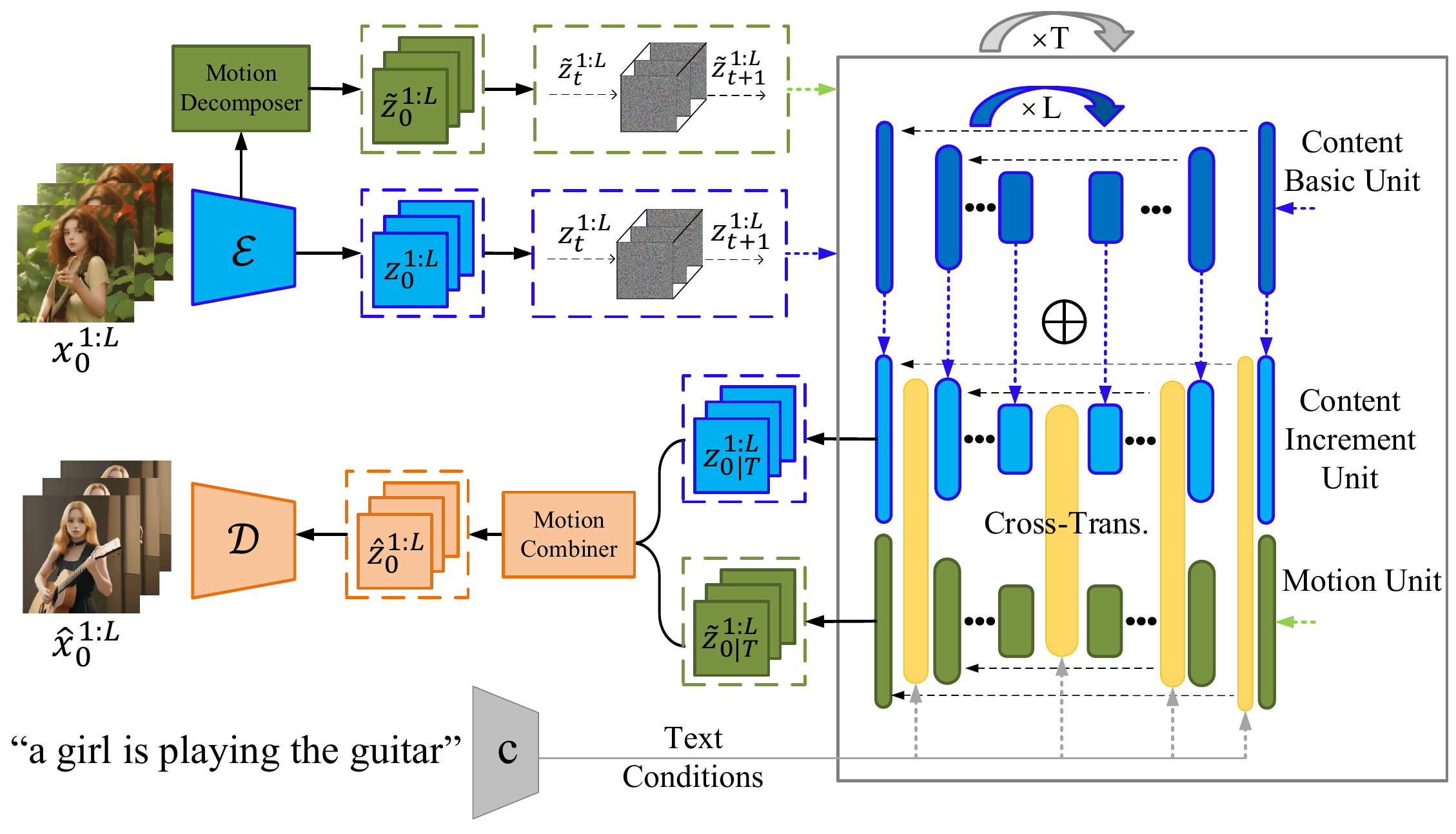}
	\caption{DSDN network framework. Initially, Content and motion features are added to noise during the diffusion process, followed by a denoising step via the dual-stream diffusion net. Lastly, the latent space features of the generated video are obtained through the motion combiner and decoded to render the final generated video.
	 }
	\label{fig:overview}
\end{figure*}

\section{Method}

In this section, we first provide an overview on the network, and then illustrate the details therein. 

\subsubsection{Overview} 
The proposed DSDN network architecture is shown in Fig.~\ref{fig:overview}. 
Initially, the input video $x_0^{1:L}$ is projected into a latent space via a frame-wise encoder $\mathcal{E}$, denoted as ${z}_0^{1:L}=\mcE({x}_0^{1:L})$, where $L$ is the length of video. Due to only the frame-wise encoding without temporal dynamics, we call ${z}_0^{1:L}$ as the content features. To mine those temporal cues for better video generation, we introduce a motion decomposer  to extract the corresponding motion information, denoted as $\tilde{z}_0^{1:L}$. Taking the both latent features as input, ${z}_0^{1:L}$ and $\tilde{z}_0^{1:L}$, we use a dual-stream diffusion way to producing personalized video content and motion variation, and subsequently propose a transformed interaction way to integrate the two components to generate a video. 

At the stage of dual-stream diffusion, two types of latent features are transformed to standard Gaussian priors through separate Forward Diffusion Processes (FDP). Then, the content feature prior undergoes denoising along Personalized Content Generation Stream (PCGS), which would result in pure denoised content features $z_{0|T}^{1:L}$. Similarly, the motion feature prior is denoised by  Personalized Motion Generation Stream, which would lead to pure denoised motion features $\tilde{z}_{0|T}^{1:L}$.  To further align the generated content and motion for suppressing flickers, we design a Dual-Stream Transformation Interaction module to bridge the two types of generation streams. After the alignment learning, we use a motion combiner to compensate dynamic information to video content, and finally form the latent feature of the target video, following by a decoder $\mathcal{D}$ to produce videos in the pixel space.

\subsubsection{Forward Diffusion Process}
To reduce resource consumption, we take the similar way to the latent diffusion model~\cite{Robin2022High}, underpinned by Denoising Diffusion Probabilistic Models (DDPM)~\cite{Jonathan2020Denoising}. Before diffusion, we use a pre-trained Vector Quantized Variational AutoEncoder (VQ-VAE)~\cite{Aaron2018Neural} to project video frames into a latent feature space, i.e., $z_0^{1:L}=\mathcal{E}(x_0^{1:L})$. For simplification, we frozen the encoder $\mcE$ during training. The content features $z_0^{1:L}$ are then processed through a motion decomposer (please see the following part: Motion Decomposition and Combination) to obtain the motion feature $\tilde{z}_0^{1:L}$. The two part of features suffer noise perturbation through a pre-defined Markov process, formally,
\begin{equation}
	\begin{split}
		&q(z_t|z_{t-1})=\mathcal{N}(z_t;\sqrt{1-\beta_t}z_{t-1},\beta_t I),\\
		&q'(\tilde{z_t}|\tilde{z}_{t-1})=\mathcal{N}(\tilde{z}_t;\sqrt{1-\beta_t}\tilde{z}_{t-1},\beta_t I),
	\end{split}
\end{equation}
where $t=1, ..., T$ and $T$ is the number of diffusion steps, $\beta_t$ defines the strength of noise at each iterative step. It is worth noting that the shared noising schedule for the two streams works well in our experience. According to DDPM, the above recursion formula could be derived into the a condensed version, 
\begin{equation}
	\begin{split}
		&z_t^{1:L} = \sqrt{\bar{\alpha_t}}z_0^{1:L}+\sqrt{1-\bar{\alpha_t}}\epsilon_1, \epsilon_1 \sim \mathcal{N}(0,I), \\
		&\tilde{z}_t^{1:L} = \sqrt{\bar{\alpha_t}}\tilde{z}_0^{1:L}+\sqrt{1-\bar{\alpha_t}}\epsilon_2, \epsilon_2 \sim \mathcal{N}(0,I), 
	\end{split}
\end{equation}
where $\bar{\alpha_t}=\prod_{i=1}^{t}\alpha_t, \alpha_t=1-\beta_t$. 
Until now, we have successfully completed the forward diffusion process for both content and motion features. This provides us with the priors $z_T^{1:L}$ and $\tilde{z}_T^{1:L}$ which are instrumental in driving the ensuing denoising process.

\subsubsection{Personalized Content Generation Stream}
In order to take advantage of well-trained image-based diffusion model, we leverage the large-scale text-to-image model, Stable Diffusion~\cite{Robin2022High}, as the fundamental model of video content generation. But to better support personalized video content generation, we design an incremental learning module to refine content generation by using a similar way to LoRA~\cite{Edward2021LoRA}. As shown in Fig.~\ref{fig:overview}, we refer to them as the content basic unit and the content increment unit, respectively. The model parameters of the two units are adaptively integrated to boost content features. Such a way not only inherits the merit of large-scale image-based generation model but also endows the creation of unique and personalized content, contributing to the overall improvement of our method.

Concretely, the content basic unit uses a modified U-Net architecture, where each resolution level incorporates 2D convolution layers with self-attention and cross-attention mechanisms. Concurrently, the content increment unit employs an extra network branch with a few tunable parameters for fine tuning. Suppose the basic unit is with parameters $W$, we have the post-tuned weight: $W'=W+\lambda \Delta W$, where $\Delta W$ is the update quantity and $\lambda$ is the step length. The hyper-parameter $\lambda$ dictates the influence exerted by the tuning process, thereby offering users extensive control over the generation outcome. To counter potential over-fitting and reduce computational overhead, $\Delta W \in \mathbb{R}^{m \times n}$ is decomposed into two low-rank matrices, as used in LoRA~\cite{Edward2021LoRA}. Let's denote $\Delta W = AB^T$, where $A \in \mathbb{R}^{m \times r}$, $B \in \mathbb{R}^{n \times r}$, and $r \ll m,n$. 

To improve the smoothness of generated content frames, we also generate video content on the condition of motion information, which refers to cross-transformer introduced in the part: Dual-Stream Transformation Interaction. Formally, we give the optimized objective on content information, 
\begin{equation}
	\mathcal{L}_{con} = \mathbb{E}_{\wtz,y,t}[\left \| \epsilon - \epsilon_\theta(z_t^{1:L},t~|~c(y),\tilde{z}_t^{1:L}) \right \|_2^2],
        \label{eq:Lcon}
\end{equation}
where $y$ is the corresponding textual description, $\epsilon_\theta(\cdot)$ here represents the part of the personalized content generation stream with the network parameter $\theta$. Note that we employ the text encoder of CLIP~\cite{Alec2021Learning} to perform the text feature extraction $c(\cdot)$.

\subsubsection{Personalized Motion Generation Stream}
In the personalized motion generation stream, we employ a 3D U-Net based diffusion model to generate a motion-coherent latent features, wherein the network architecture of 3D U-Net is similar to that in \cite{Jonathan2022Video}. The reason why use 3D U-Net is that the global motion variation of the entire input video could be captured for subsequent motion generation. Given an input sequence of motion priors $\wtz_{T}^{1:L}$, we can obtain a transformed representation vector after using the encoding stage of 3D U-Net, and the next denosing process takes the vector as input for diffusion, similar to DDPM. Differently, to make the generated motion matched with content, we use the generated content feature $z_t^{1:L}$ as well as the text prompt $c(y)$, as the conditions, in the denoising diffusion process. The use of content condition refers to cross-transformer, which will be introduced in the next part: Dual-Stream Transformation Interaction. Hereby, the training objective of the personalized motion generation stream can be formulated as:
\begin{equation}
	\mathcal{L}_{mot} = \mathbb{E}_{z, y, t}[\left \| \epsilon-\epsilon_{\tilde{\theta}}(\tilde{z}_{t}^{1:L},t~|~c(y),z_{t}^{1:L}) \right \|_2^2],
        \label{eq:Lmot}
\end{equation}
where $\epsilon_{\tilde{\theta }(\cdot)}$ represents the part of the personalized motion generation stream with the network parameter $\tilde{\theta}$.

\subsubsection{Dual-Stream Transformation Interaction} 

To well align the generated content and motion information, we design a cross-transformer interaction way between the two denoising streams. On the one hand, we infuse the denoising generation procedure of the motion stream with conditional feature information from the content, by taking the transformer from content to motion,  which would enhance the continuity of the overall motion. On the other hand, the denoising generation process of the content stream also absorbs the conditional feature information from the motion, by taking the transformer from motion to content. This cross-transformer based streams render the overall content smoother, creating a synergistic effect that enhances the consistency and quality of the final output.
In detail, after each convolutional layer of U-Net, we interpose a cross-attention layer to integrate the latent features of both content stream and motion stream. Taking the case from motion to content as an example, formally, we have
\begin{align}
z_{con} \!\!=\!\! \text{Att}(Q_{mot}, K_{con}, V_{con}) \!\!=\!\! \text{Softmax}(\frac{Q_{mot}K_{con}^T}{\sqrt{d}}) \cdot V_{con},
\end{align}
where $Q_{mot}=W^Qz_{mot},K_{con}=W^Kz_{con},$ and $V_{con}=W^Vz_{con}$ denote three projections of cross-attention along the content stream with the parameters $W^Q, W^K, W^V$, and $d$ is the feature dimensionality. The motion stream features can constrain the content stream generated to ensure smoother transitions from frame to frame. Similarly, the same principle applies for the motion stream as well. At this time, the content stream features can supply an understanding of image apparent information for the generation of motion latent features during the denoising process. Hence, the cross-attention layer in this context facilitates mutual conditioning between the dual-stream features. 

\begin{figure}[!htp]
	\centering
	\includegraphics[width=0.5\textwidth]{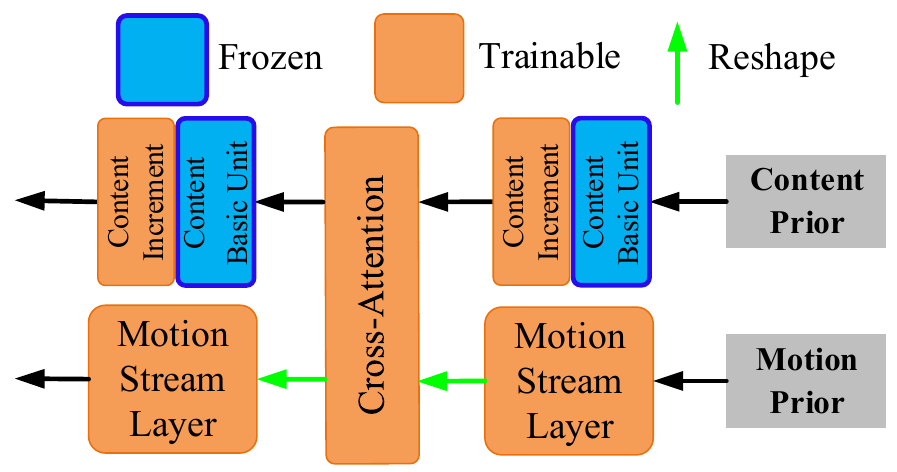}
	\caption{Dual-stream transformation block. 
	 }
	\label{fig:suboverview}
\end{figure}

Intuitively, such a cross-transformer strategy is explicitly performed on two branches of diffusion processes. This is very different from those previous video diffusion methods~\cite{Andreas2023Align, Yuwei2023AnimateDiff, Wenyi2022CogVideo}, which essentially use a single-stream diffusion process by either directly inserting a pseudo 3D layer to manage temporal information, or intercalating a 3D layer between two successive 2D convolution layers. Besides, the dual-stream diffusion network also take the corresponding textual conditional embedding as input. 

In the end, the total optimization objective of dual-stream diffusion net is the joint in both Eq.~\ref{eq:Lcon} and Eq.~\ref{eq:Lmot}.
Throughout the optimization process, only the content increment unit (in the content stream) and the cross-attention layer (between two denoising streams) are trainable, whilst the content basic unit, i.e., the underlying text-to-image model in the content stream, remains static in order to preserve the consistency of its feature space. An illustration is shown in Fig.~\ref{fig:suboverview}.

\begin{figure}[!h]
	\centering
	\includegraphics[width=0.50\textwidth]{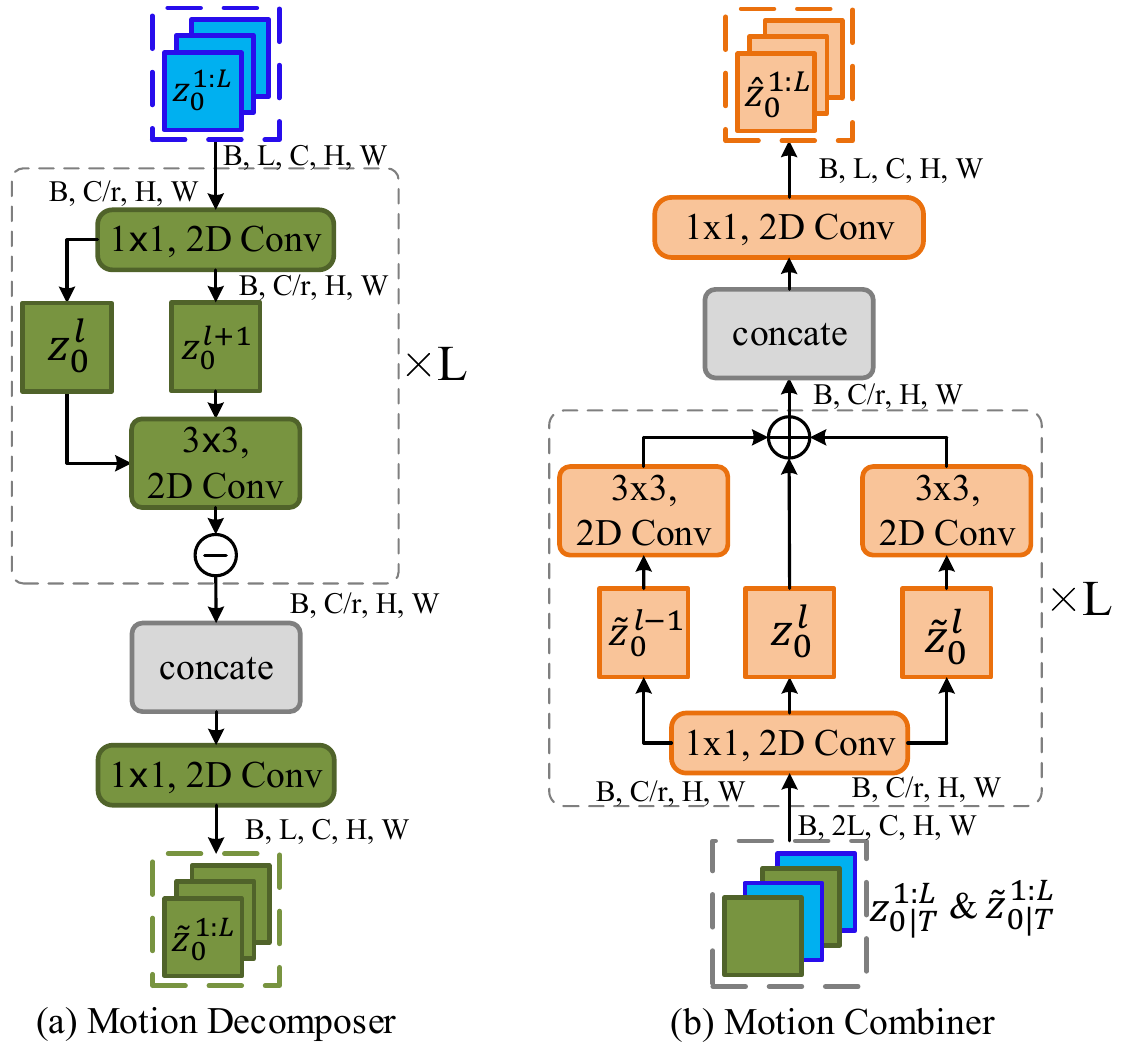}
	\caption{Details of Motion Decomposer and Motion Combiner.
	 }
	\label{fig:motionGC}
\end{figure}

\subsubsection{Motion Decomposition and Combination}

In order to separate motion features and reduce computational cost, we design the lightweight motion decomposer and corresponding motion combiner inspired by the work~\cite{Boyuan2019STM}. 
In terms of the motion decomposer, given the input content feature $z_0^{1:L} \in \mathbb{R}^{B\times L \times C \times H \times W}$, the motion decomposer first utilizes a $1 \times 1$ convolution layer to reduce the channel ($C$) by a factor of $r$, which would alleviate computing expense. We then compute motion features derived from every pair of sequential frames. For instance, given transformed $z_0^l$ and $z_0^{l+1}$, we initially apply a 2D channel-wise convolution to $z_0^{l+1}$. Subsequently, we subtract this new value from $z_0^{l}$ to obtain the $(l)$-th and $(l+1)$-th frame motion representation $\tilde{z}_0^{l}$, formally, 
\begin{equation}
	\tilde{z}_0^{l} = conv(z_0^{l+1}) - z_0^l,
\end{equation}
where $conv(\cdot)$ denotes a 2D channel-wise convolution. 
As depicted in Fig.~\ref{fig:motionGC}(a), we apply the motion decomposer on every two adjacent frames. As a result, the motion decomposer generates $L-1$ frames of motion features. To ensure compatibility with the original temporal length, we append the last frame of video to reach the length $L$ in our experiment. Finally, another $1\times 1$ 2D convolution layer is utilized to restore the number of channels back to $C$.

For the motion combiner, given the denoised content and motion features $z^{1:L}$ and $\wtz^{1:L}$, we also first employ $1\times 1$ convolution to reduce the channel number. As shown in Fig.~\ref{fig:motionGC}(b), the content feature and their adjacent motion features are fused after doing a 2D convolution on motion features. Formally, the $l$-th frame latent feature $\hat{z}_0^{l}$ of the generated video is defined as,
\begin{equation}
	\hat{z}_0^{l} = conv(\tilde{z}_0^{l-1}) + z_0^l + conv(\tilde{z}_0^{l}),
\end{equation}
where $conv$ represents the 2D channel-wise convolution. Upon acquiring the combined video latent features, it comes back to the original channel dimension via a $1\times 1$ convolutional layer. This combined features are then input into the final decoder, which yields a video in the pixel space.

\begin{figure}[!thp]
	\centering
	\includegraphics[width=0.50\textwidth]{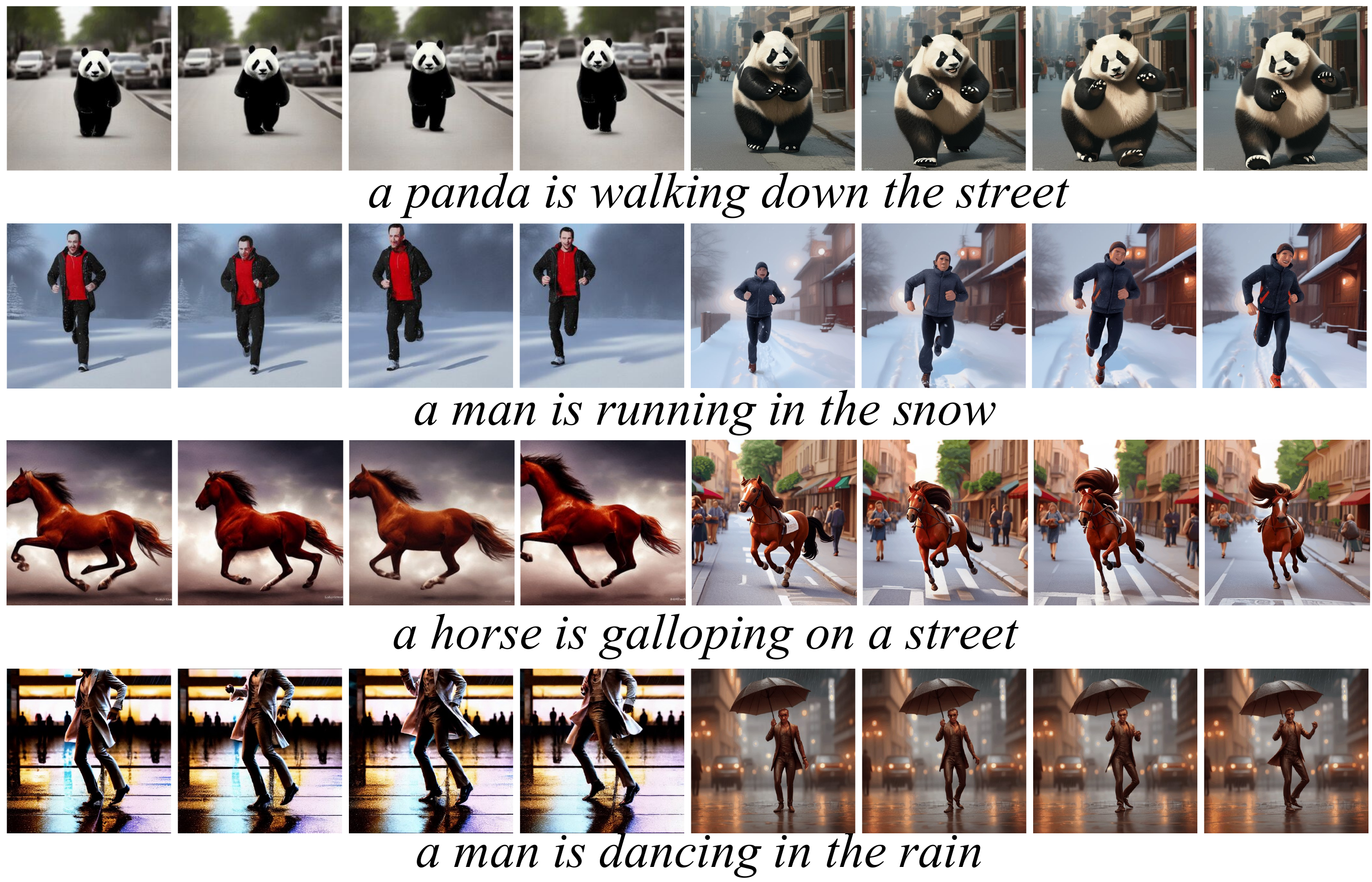}
	\caption{ Qualitative comparison between Text2Video-Zero~\cite{Levon2023Zero} (frames 1-4 in each row) and our method (frames 5-8 in each row). Please see the videos in the website.
	 }
	\label{fig:compare}
\end{figure}

\section{Experiments}
\subsection{Implementation Details}
In our experimental setup, we generate $L=16$ frames with a resolution $512\time 512$ for each video. We trained the Dual-Stream Diffusion Net using a subset (comprising 5M videos) from the WebVid-10M~\cite{Max2021Frozen} and HD-VILA-100M~\cite{Hongwei2022Advancing} datasets. Video clips within the dataset are first sampled at a stride of 4, then resized and centrally cropped to a resolution of $256\times 256$. For the content basic unit, we employ stable diffusion v1.5 pre-trained weights which remain frozen throughout the training procedure. The content incremental unit does not commence training from scratch, but rather utilizes an existing model~\cite{Civitai2022} as pre-trained weights, followed by fine-tuning on our training data. For the motion unit, we initialize our Personalized Motion Generation Stream with the weights of LDM~\cite{Robin2022High}, which were pre-trained on Laion-5B~\cite{Christoph2022LAION}. During inference, it takes approximately 35 seconds to sample a single video using one NVIDIA RTX 4090 GPU.

\subsection{Comparison with Baselines}
We compare our method with two publicly available baseline: 1) CogVideo~\cite{Wenyi2022CogVideo}: a Text-to-Video model trained on a dataset of 5.4 million captioned videos, and is capable of generating videos directly from text prompts in a zero-shot manner.
2) Text2Video-Zero~\cite{Levon2023Zero}: also a Text-to-Video generation method based on the Stable Diffusion model. Since our method is a text-to-video method we compare with Text2Video-Zero in pure text-guided video synthesis settings.
Owing to space constraints, we present a quantitative comparison of our method with the two aforementioned approaches. However, for qualitative results, we limit our comparative analysis to the superior performing text2video-zero method for a more focused and meaningful evaluation.

\begin{table}
	\centering
	\caption{Comparison of CLIP score metric with baselines.}
	\scriptsize{
		\setlength{\tabcolsep}{10pt}{
			\begin{tabular}{clccccccccccc}
				\toprule
				&Methods                  &Frame Consistency &Textual Alignment\\
				\toprule
				&CogVideo                 &88.32             &22.02 \\
				&Text2Video-Zero          &90.21             &29.56 \\
				&Ours  			            &\textbf{92.13}    &\textbf{32.23} \\
				\bottomrule
			\end{tabular}
		}
	}
	\label{table:comparison}
\end{table}

\begin{figure}[!htp]
	\centering
	\includegraphics[width=0.50\textwidth]{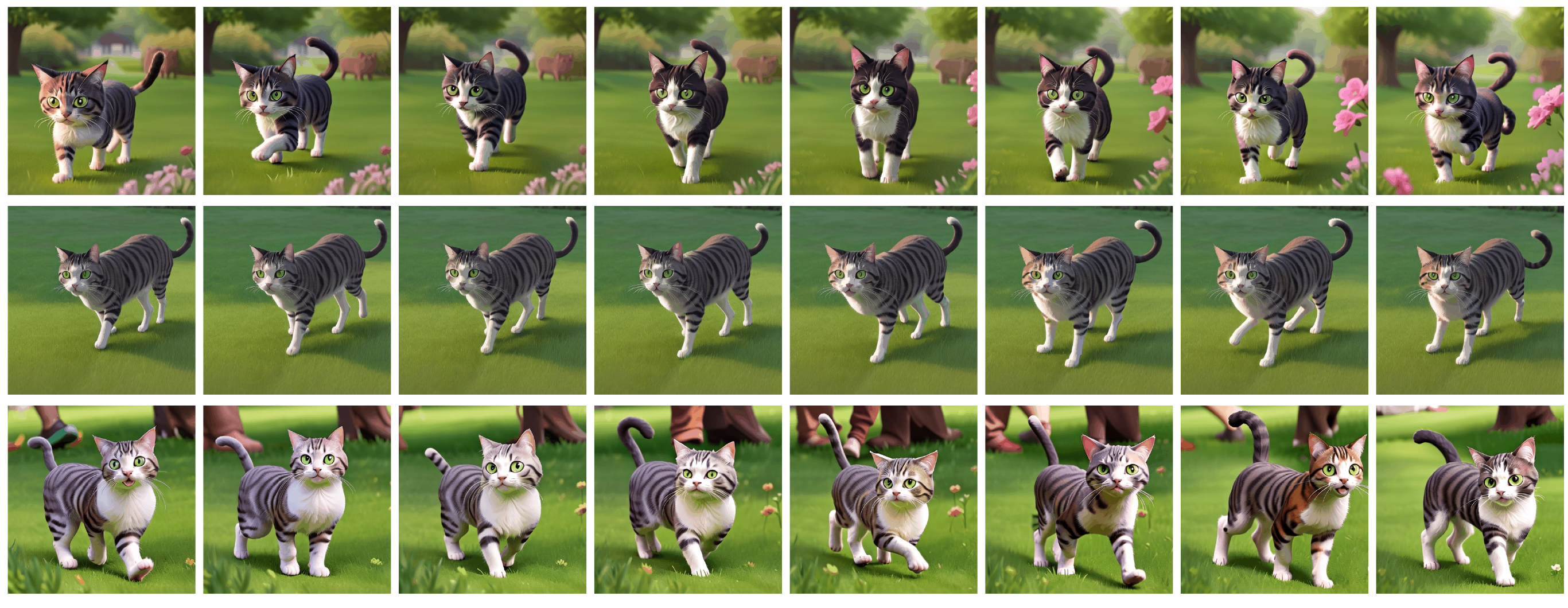}
	\caption{The diversity of our method qualitative results. Prompt: a cat is walking on the grass.
	 }
	\label{fig:diversity}
\end{figure}

\begin{figure*}[!htp]
	\centering
	\includegraphics[width=0.95\textwidth]{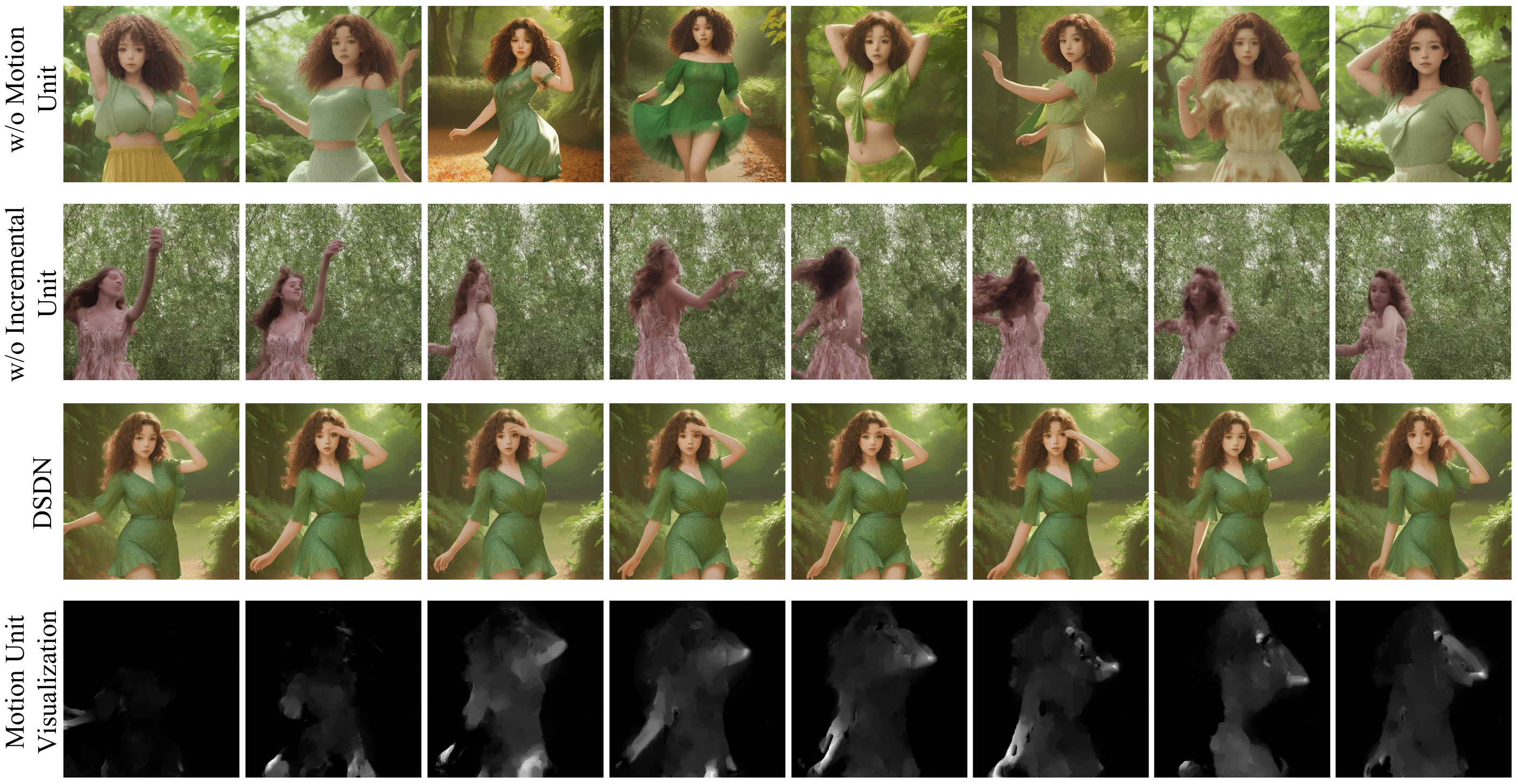}
	\caption{Ablation study. Prompt: a girl is dancing among leaves, curly hair.
	 }
	\label{fig:ablation}
\end{figure*}

\subsubsection{Quantitative Comparison}
We evaluate our method in relation to baseline models by employing automated metrics, detailing frame consistency and textual alignment outcomes in the accompanying Table.~\ref{table:comparison}. For assessing frame consistency, we compute CLIP~\cite{Alec2021Learning} image embeddings on all frames within the output videos, reporting the average cosine similarity across all pairings of video frames.
To evaluate textual alignment, we calculate the average CLIP score between all frames in the output videos and their corresponding prompts. Our findings reveal that the videos generated via our method surpass publicly accessible alternatives such as CogVideo~\cite{Wenyi2022CogVideo} and Text2Video-Zero~\cite{Levon2023Zero}, particularly with regard to frame consistency and text alignment. This suggests that our method offers a more robust and coherent approach to video generation from textual prompts.

\subsubsection{Qualitative Comparison}
When compared to text2video-zero, our method demonstrates superior consistency in both content and motion across generated videos, as shown in Fig.~\ref{fig:compare}. As illustrated in the first row, observable body swings are evident as the panda walks, while in the second row, we see the limbs swing as the person runs, accompanied by gradual changes in the background. In the third row, the lower limbs of the horse are seen swinging as it gallops, set against a dynamic background.
Furthermore, our method outperforms the other approach in terms of content quality and its conformity with the text. For instance, the generated pandas in our model appear more realistically rendered, the snow in the second row exhibits imprints, and the street in the third row is more logically constructed.
We further include a fourth row as a comparative example in a less favourable environment - rain conditions. Here, the content generated by the other method appears unrealistic, whereas our method not only captures the essence of a rainy day more effectively but also establishes the logical connection between rain and umbrellas, thereby enhancing the realism and context-appropriateness of generated videos.

Beyond this, we further show the qualitative results of our method on video diversity generation, as shown in Fig.~\ref{fig:diversity}. 
Using "a cat is walking on the grass" as the text input, we can see various actions such as a cat walking forward, left, and right. Impressively, the generated videos also exhibit diversity in aspects like fur color, body shape, and pose, thereby encompassing a rich variety in content.
Concurrently, the generated video preserves high continuity. As demonstrated in the first row, the flower at the lower right gradually comes into view, while in the second row, subtle changes in the cat's shadow can be discerned. In the third row, a figure in the background is progressively moving, further enhancing the sense of dynamic realism.
Furthermore, we acknowledge a minor failure case depicted at the end of the third row, where the color of the cat appears slightly altered. This issue primarily stems from the generated background inadvertently influencing the content creation process itself. However, as evident from the comprehensive results showcased, such instances are rare, thereby attesting to the robustness and reliability of our proposed method in text-to-video generation.

\subsection{Ablation Study}
We conduct a rigorous ablation study to evaluate the significance of both the content increment unit and the motion unit, as depicted in Fig~\ref{fig:ablation}. Each design component is selectively ablated to determine its individual impact on the model's overall performance.

\subsubsection{Motion Unit}
The outcomes from the first row indicate that while a model void of the motion unit can still synthesize apparent content in accordance with text conditions, it fails to maintain the continuity between video frames. This result stems from the absence of temporal dimension modeling—resulting in generated video frames being independently constrained by text conditions without inter-frame connection. In terms of content congruity, the generated video frames exhibit a solid alignment with the narrative conveyed by the textual conditions. For instance, elements like 'dancing', 'leaves', and 'curly hair' described in the text are accurately manifested within the generated imagery. 

\subsubsection{Incremental Unit}
As observed in the second row, the visible content quality suffers a significant reduction without the incremental unit model. This underscores the pivotal role of the content increment unit in learning richer visual content beyond what the content base unit alone can achieve.
Upon analyzing the results from the first three rows, we observe a few issues: 1) Fine-tuning the incremental unit seems to stabilize the apparent content; for instance, the girls in both the first and third rows face forward, whereas without the incremental unit, as seen in the second row, the girl's perspective can emerge from any direction.
2)The clothing color in the first and third rows leans towards green, mirroring the hue of the background environment. These challenges might arise due to limited parameter volume within the incremental unit, thereby restricting the scope of apparent content it can learn effectively. Such observations underscore areas for further exploration and improvement in the incremental unit of our method.

\subsubsection{Motion Unit Visualization}
Furthermore, we offer a detailed visualization of the motion unit’s effects of the third row in the last row. The visualizations highlight the efficacy of the motion unit in accurately capturing inter-frame motion details such as arm swings, body movements, and hair fluttering, thereby underscoring its critical role in achieving a coherent and dynamic video output.

\section{Conclusion}
This work presented a novel dual-stream diffusion net (DSDN) to improve the consistency of content variations in generating videos. Specifically, the designed two diffusion streams, video content and motion branches, could not only run separately in their private spaces for producing personalized video variations as well as content, but also be well-aligned between the content and motion domains through leveraging our designed cross-transformer interaction module, which would benefit the smoothness of generated videos and enhance the consistency and diversity of generated frames, where the motion is specifically modeled as a single branch that distinguishes from most existing video diffusion methods. Besides, we also introduced motion decomposer and combiner to faciliate the operation on video motion. Qualitative and quantitative experiments demonstrated that our method produces better continuous videos with fewer flickers.

\bigskip

\bibliography{aaai24}

\end{document}